# Collapsing the Decision Tree: the Concurrent Data Predictor

**Cristian Alb**

*ca.publicus@gmail.com*

### Abstract

A family of concurrent data predictors is derived from the decision tree classifier by removing the limitation of sequentially evaluating attributes. By evaluating attributes concurrently, the decision tree collapses into a flat structure. Experiments indicate improvements of the prediction accuracy.

**Keywords:**
Machine Learning; Supervised Classification; Decision Trees

## 1    Introduction

Techniques in machine learning are used to discover correlations, patterns, and trends in data. They also classify, or predict, data outcomes based on knowledge of a training data set. A great variety of algorithms exist and many combinations thereof are possible [1, 2, 3].

Restricting the area of interest to supervised learning, the most widely used families of algorithms are:

1. Decision trees
2. Random forests
3. Neural networks
4. Naive Bayes
5. K-nearest neighbor
6. Support-vector machines
7. Linear regression
8. Logistic regression
9. Linear discriminant analysis

By design, algorithms 5-9 are meant to be used with data sets that have numerical attributes. Algorithms 1-4 do operate with categorical attribute data sets.

A "Concurrent Data Predictor" family of algorithms is proposed here. These algorithms are designed to operate with categorical attributes. Unlike the aforementioned algorithms, the Concurrent Data Predictors do not require a training phase.

Decision trees and random forests proved to be quite successful in a variety of classification and prediction tasks [4, 5]. Random Forest classifiers extend the operation of decision trees by means of ensemble methods.

However, decision trees might not fully exploit the informative content of the training data; they evaluate attributes sequentially and ignore possible synergies that arise when evaluated concurrently.

## 2 Decision Tree Operation

Decision tree operation is based on the assumption that attribute values better predict the target outcomes when evaluated in a specific order determined during the training phase [6].

The training data set is equivalent to a table of values. Each row can be thought of as a training instance, or entry. Columns correspond to attributes, or features, associated to the training data set. The target column is a particular column representing the outcome of interest. It classifies the row instance according to the outcome value. The other attribute columns contain values that characterize each instance from the perspective of the known features.

In Fig. 1, a simple example illustrates a training data set and its associated decision tree. The trapezoids represent the leaves and they contain the associated target outcomes.

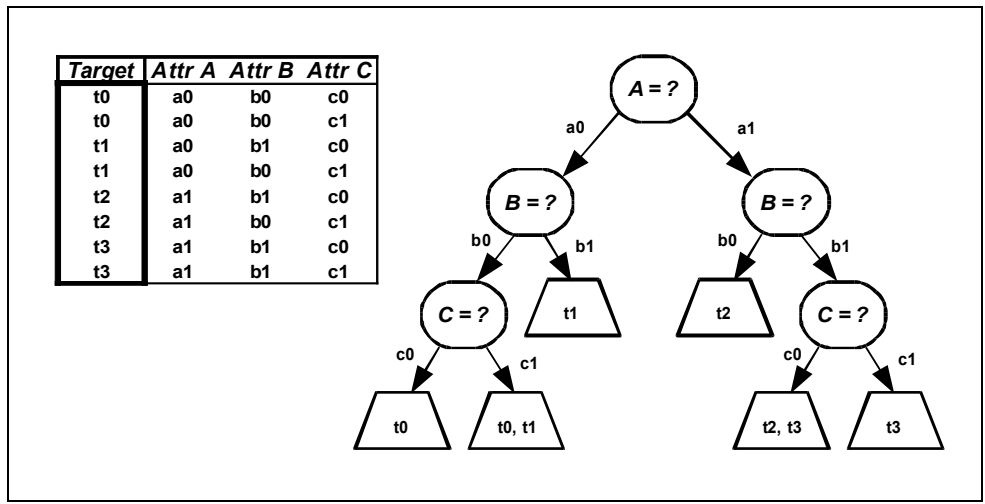

Figure 1: Decision tree example.

The goal of the classifier is to predict the most likely target outcome for a given query entry. The query entry consists in a list of values that correspond to the column attributes of the training data set. The decision tree algorithm starts at the root with the set of all target outcomes in the training data set. Each of the column attributes is evaluated to determine how it partitions the target outcomes. An impurity measure is used to determine which attribute best partitions the target outcomes. Ideally, for each attribute value, the corresponding outcome subset contains identical target values.

Fig. 2 illustrates an example for the evaluation of the most suitable attribute for splitting the decision tree. Entropy is used as the impurity measure; attribute A is the best choice while attribute C is the worst.

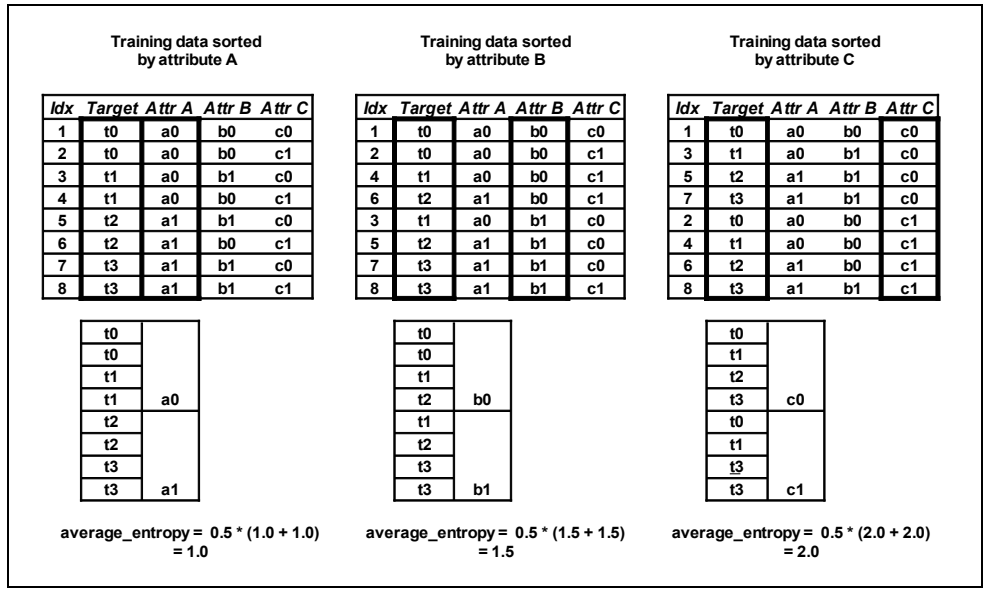

Figure 2: Decision tree attribute selection.

Each subset of target outcomes becomes a new node branching out from the parent node. The process is recursively repeated on each of the branched outcome subsets. The process stops when no more attributes are left, or the current target outcome set is perfectly homogeneous. Such terminal node, or leaf, is used to produce the outcome prediction corresponding to the attribute values of the branch. A majority voting is performed on the node's subset of outcomes in order to elect the predicted outcome.

The impurity measure, used as a splitting criteria, is an average of the homogeneity of the split subsets of target outcomes. The homogeneity of the outcomes is usually expressed by the entropy of the ensemble. Other measures can be used, as is the Gini index, or information energy [7].

# 3    Splitting Outcomes with Ensembles of Attributes

Decision trees split the current set of outcomes using the values of one selected attribute. The selected attribute is the one that achieves the best score using the splitting criteria.

What happens if this attribute is chosen at random? The resulting algorithm still retains a predictive power as experimental results indicate. Such a decision tree will be referred to as a "Random Tree".

Why use only one attribute instead of multiple ones? It is conceivable that the split subsets that result from evaluating more than one attribute generate a better impurity measure.

Fig. 3 depicts an example of a classic decision tree that uses one attribute at a time to split the target outcomes corresponding to a node.

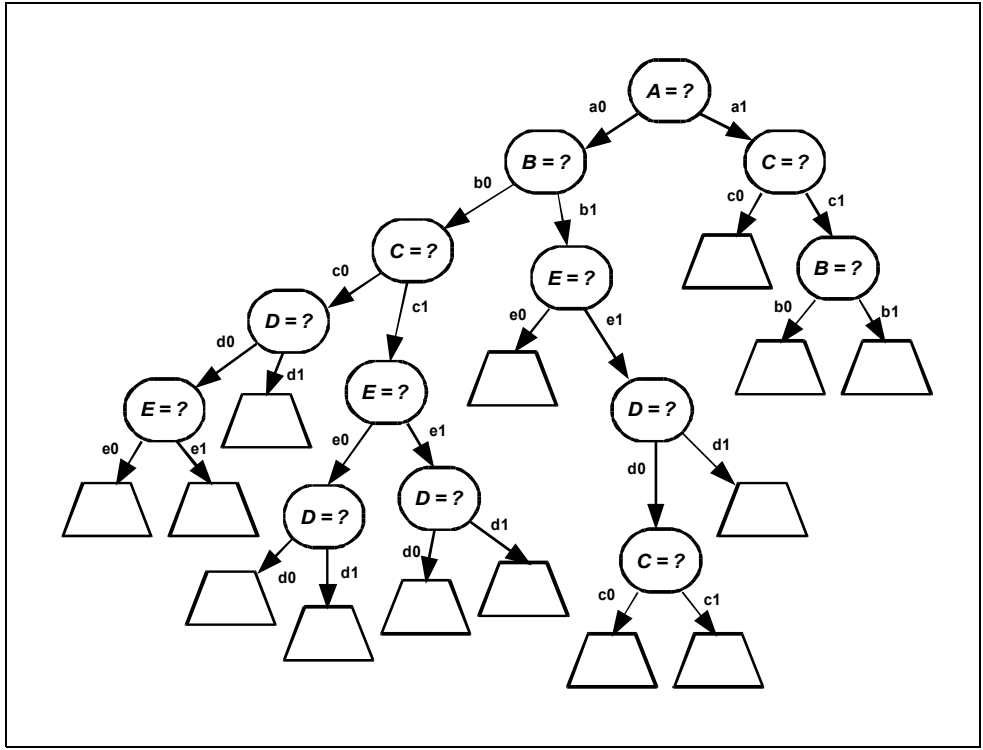

Figure 3: Decision tree, max depth 5.

Fig. 4 depicts a decision tree where two attributes are used to split the root node. To be noted how the evaluation of the impurity of all possible combinations leads to a combinatorial explosion of branches. The same phenomenon affects the choice of the best combination of attributes.

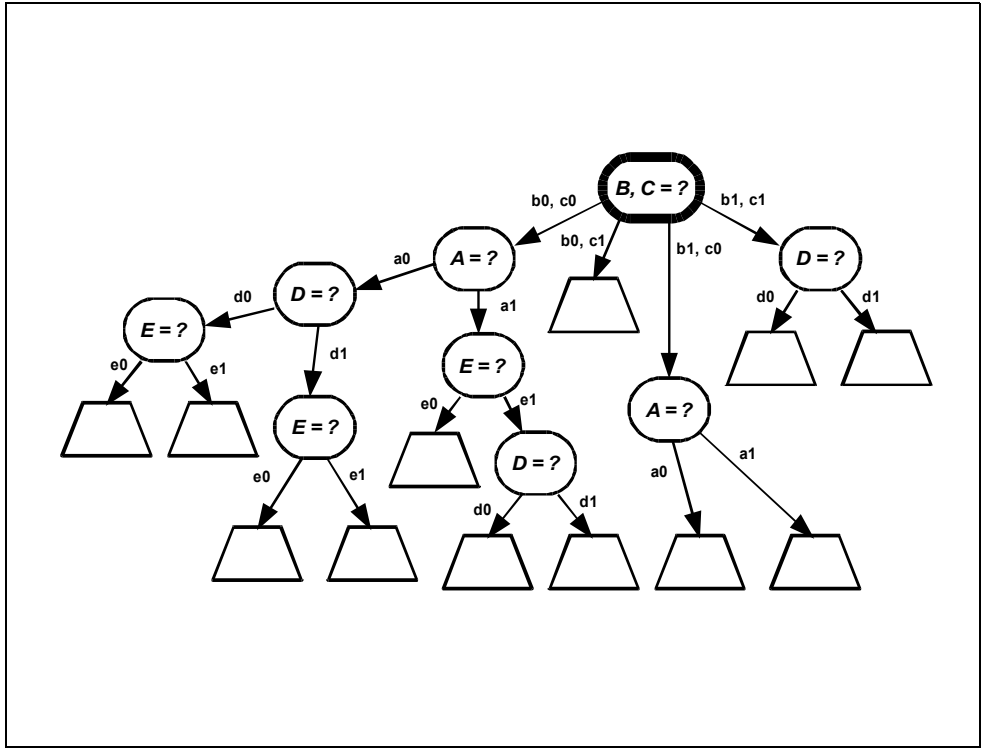

Figure 4: Decision tree, max depth 4.

Fig. 5 depicts a decision tree where two attributes are used for splitting more nodes. The principle could be extended to the evaluation of three, four, five, etc., attributes per node. If all the attributes are used at the same time, the tree collapses into a flat structure.

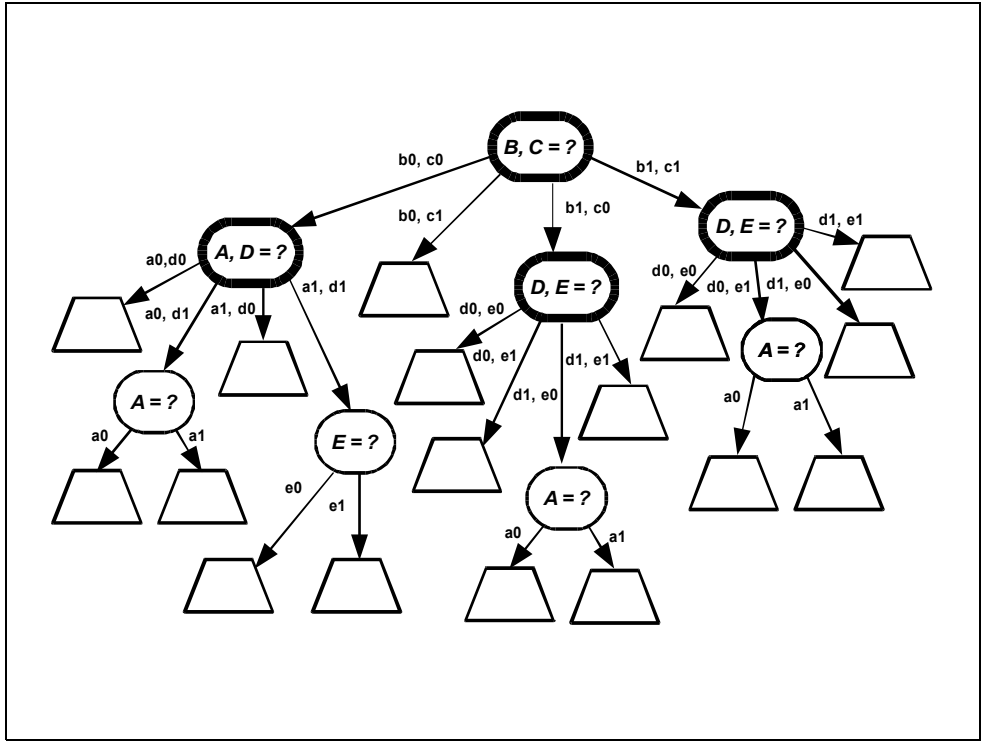

Figure 5: Decision tree, max depth 3.

Fig. 6 depicts a decision tree where all attributes are used for splitting. The collapsed tree has a uniform depth of one node.

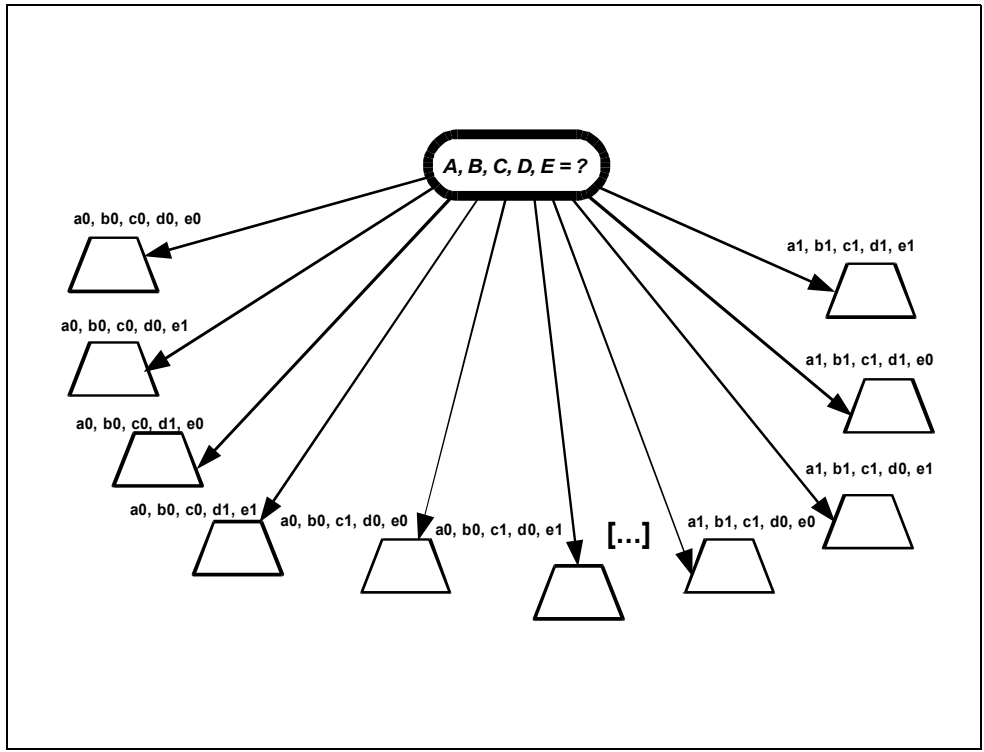

Figure 6: Decision tree, max depth 1.

A consequence of this collapse is the simplification of the processing; it is no longer necessary to evaluate and compare the impurity of the splits, there is only one exhaustive split left.

# 4    Concurrent Data Predictors    *(deodata)*

The above technique enables a different type of classifiers, the "Concurrent Data Predictors", also referred to as "*Deodata*" methods.
The result of the tree collapse is a flat structure, a collection of subsets of target outcomes corresponding to each attribute combination present in the training data set.

In general, a large number of attribute value combinations are present in the training data. This does not warrant the effort of precalculating the predictor outcome as part of a preprocessing phase. The prediction can be obtained by parsing each training entry and comparing it to the query. Such a mode of operation is referred to as "lazy learning", also "just-in-time learning". However, for specific applications, the target outcomes could be grouped together by attribute value combination. Such a preprocessing phase enables a faster operation when predicting queries.

### Conceptual Operation

An unknown query entry is evaluated by comparing its combination of attribute values with the corresponding attribute values in each of the training entries in the data set.

Each attribute value of the currently evaluated training row will be compared with the corresponding attribute value of the query entry. A match column score is calculated for each attribute. The match column score can be, in the simplest form, a value of one for a match and a zero otherwise. Once all attributes have been compared, an entry match score is computed for the currently evaluated training row. In the simplest form, this score is the sum of the match column scores of all attributes.

Conceptually, a row summary ensemble is generated for the currently evaluated training row. The summary includes the target outcome and the entry match score. The row summary ensemble will be used to update the contents of a work data set. This work data set aggregates the content of the row summary ensemble for each of the evaluated rows of the training data set.

After the parsing of the rows is completed, the work data set will be processed in order to estimate likelihood measures for the target outcomes. For prediction or classification tasks, the target outcome with the highest likelihood will be used as output.

## 5 Proximity *(delanga)*

Classification, or prediction, can be done in a way that closely resembles the way decision trees operate. The prediction is given by the majority voting in the subset of outcome values that best matches the attribute values of the query entry. This type of method is referred to as a "Proximity" variant of the Concurrent Data Predictors; it is also referred to as a "Delanga" method.

### Conceptual Operation

The Proximity methods create a work data set that consists of an ensemble of target outcome collections. These collections can be viewed as lists, where the actual order of the elements is not relevant.

Each of these target outcome collections will be associated to an entry match score. These collections will be referred to as match score lists.

After the row summary ensemble is generated for a training row, the match score list corresponding to the entry match score will be updated by appending the target outcome to the list. If a list does not exist for the entry match score, it will be created and will contain the target outcome as element.

After the parsing of the rows is completed, the match score list corresponding to the top score is selected. The counts of the target outcomes will constitute the likelihood measures.

A tie situation occurs when counts are equal for a group of target outcomes. For these situations, a tie breaking procedure can be used. A tie breaking procedure consists in evaluating the next best match score list. If a tie still persists, the next best match score list is evaluated, etc.

# 6    Cascading    *(varsate)*

The "Proximity" variant does not require the computation of impurity measures like entropy or Gini index. A variant that aims to achieve a more robust prediction and does use such measures is described next. This type of method is referred to as a "Cascading" variant of the Concurrent Data Predictors; it is also referred to as a "Varsate" method.

## *Conceptual Operation*

The Cascading methods extend the operation of the Proximity methods by aggregating the work data set into a set of cascaded match lists.

The processing is identical with that of the above described Proximity method, up to the generation of the match score lists. Once the lists are generated, it is required to order them such that the list with the best score is placed on top. The match score list at the top becomes the top cascaded match list. This top list will be appended to the next match score list of the work data set; the result is the cascaded match list for the lower level. The content of this lower level cascaded match list will be appended to the next lower level match score list, etc. The process continues until all the lists have been processed.

Each of the resulting cascaded match lists will be evaluated and assigned a predictive score. The predictive score could be an impurity measure such as entropy, Gini index, or some other information content measure. The list with the best predictive score will be used to derive the likelihood measures for classification, prediction, or data analysis. For instance, in case of prediction, the target outcome with most counts in the selected list becomes the predicted outcome. Again, a tie breaking procedure can be used in situations where the target outcome counts are equal.

# 7    Swapped    *(rasturnat)*

The "Swapped" variant results from "Proximity" by swapping the role of the entry match score with that of the target outcome. In the work data set, instead of aggregating target outcomes to a match score ensemble, the method aggregates the match score to the corresponding target outcome.

There are many ways to update the target outcome score. A simple implementation consists in counting the number of attribute matches into an entry match score, using that as an argument to an exponential function, and then adding the resulting value to the target outcome score. A more complex scoring procedure could make use of additional parameters like:

- available number of entries in the training data set
- number and variety of target outcomes
- number and importance of attributes in the training data set
- distribution and other characteristics of the attribute's values

This type of method is referred to as a "Swapped" variant of the Concurrent Data Predictors; it is also referred to as a "*Rasturnat*" method.

### *Conceptual Operation*

The Swapped methods create a work data set that consists of an ensemble of target outcome entries, where each entry has a cumulative outcome score.

After the row summary ensemble is generated for a training row, the corresponding target outcome score entry of the work data set is updated with the entry match score. The updating could consist in applying a transformation function to the entry match score and adding the resulting value to the outcome score. The transformation function could be an exponential function like a power of two, or a Gaussian.

After the parsing of the rows is completed, the work data set will contain the target outcome likelihood measures.

Code excerpts illustrate the operation of these methods in Appendix A.

## 8      Algorithm Converging Equivalence

The following three algorithms are considered: the Proximity Concurrent Data Predictor, the ID3 Decision Tree, and the Random Tree. They operate differently and, in general, will yield different predictions over multiple predictions tests. However, for a very large training data set, their predictions will converge; they will predict the same target outcomes.

A rationale for this convergence is the following: for a very large training set, all possible combinations of attribute values will be present. As a consequence of this data saturation, the number of outcomes corresponding to any given combination of attributes will also be large. The shape of the bins of outcome counts will approach the intrinsic probability distribution of outcomes for that particular attribute combination. The decision tree algorithms will have as leaves such distributions of outcomes. No matter the ordering chosen for the attribute nodes in the tree, the large number of training entries will provide enough outcomes for each attribute combination in the leaves.

The same reasoning applies to the Proximity Concurrent Data Predictor that specifically accumulates training outcomes into the best match score list. For a very large training set, the best match score list will correspond to the perfect match of attributes. In a theoretical sense, for a very large training data set, the three algorithms are equivalent, approaching the saturation/ideal accuracy.

A diagram of such a converging evolution is shown in Fig. 7

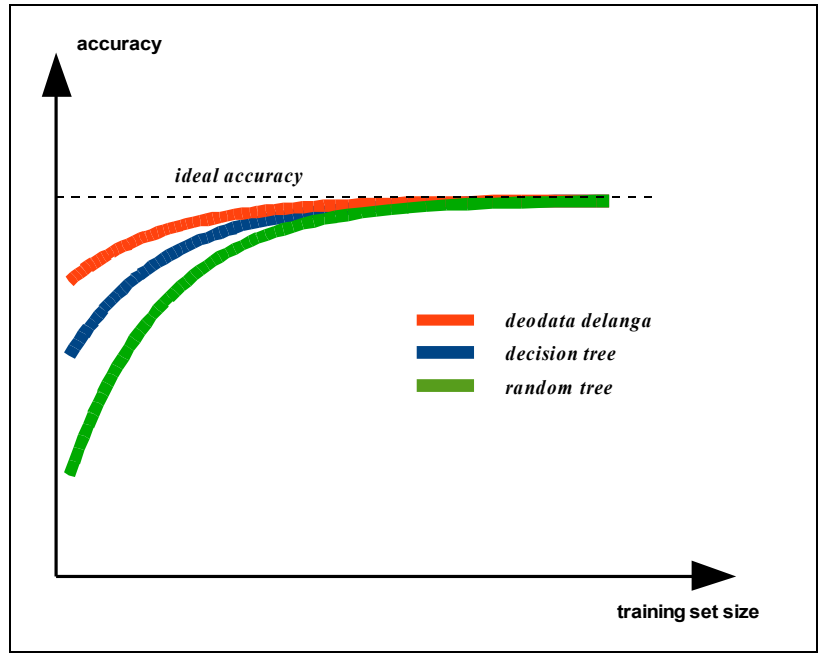

Figure 7: Converging accuracy.

Experimental results do show such a converging tendency. The accuracy of the algorithms increases with the number of samples while the differences among them diminish.

# 9  Comparison with k-Nearest Neighbor

The k-Nearest Neighbor (k-NN) classifier is an algorithm that relies on a distance metric for classification [8]. It was intended to be used with attributes that represent continuous (numerical) variables.

The operation mode of the k-NN classifier resembles the lazy learning implementation of the concurrent data predictors. In particular, the Proximity variant of these predictors is comparable to a "discretized" version of the k-NN classifier. If a Hamming distance is used for matching attributes, the resulting k-NN classifier is similar to the Proximity variant.

The difference consists in how the group of neighbors is selected for voting. The k-NN classifiers use "k" as a parameter to specify the size of the group. The Proximity variant does not impose such a number, using instead the number of entries that qualify as "nearest". Could be one or many.

The Proximity methods could be described as "nearest neighborhood", where the neighborhood contains one or more predictive entries. The "next closest neighborhood" can be used to break ties when several outcomes have the same number of occurrences in the "nearest neighborhood".

What would be a "continuous" counterpart to the "discrete" Concurrent Data Predictors described so far?

Would the continuous counterpart of the Proximity variant be more similar to the k-NN?

No, another distinction arises in how the proposed methods evaluate closeness. For each attribute/column, the Proximity variant evaluates the degree of match between the corresponding values in the query and training entries and subsequently aggregates them. The k-NN algorithms evaluate closeness, or proximity, with a distance metric that operates on a pair of attribute value vectors.

# 10    Continuous Concurrent Data Predictors

Here, the focus has been on classifiers for categorical attributes. But, as mentioned above, continuous implementations of the Concurrent Data Predictors are possible. A continuous implementation uses attributes that represent continuous (numerical) variables. As such, the match column score cannot be a binary value anymore; a continuous number is more adequate.

One possibility is to normalize the attribute values of the column and estimate the distance between the two compared values as a function of the standard deviation separating them.

# 11    Experimental Results

A set of tests have been performed in order to compare the prediction accuracy of the proposed methods.

The testing was done using a data set from the scikit-learn package "UCI ML hand-written digits". The data set consists in 8x8 gray-scale images of hand-written digits. The data sets used in testing were derived from the original through several alterations:

- scaling the 8x8 resolution to lower resolutions, e.g., 6x6, 4x4, 3x3, etc.
- selecting only a few random pixels from the scaled digit image
- resampling the pixels gray-scale intensity to a lower resolution
- restricting the range of possible outcomes to a set of selected digits

These alterations provide a synthetic diversification of the data set.

Many combinations have been tried and the results appear to be relatively similar. The results presented in Table 1 are for a prediction test with four possible target outcomes (digits) using six attributes (random pixels) with a variable number of attribute values (pixel intensities). Note that the pixel intensities were treated as categorical/nominal values and not as numerical ones. The number of entries used in the training data set was 24 (6*4, six examples for each target digit).

The test consisted in predicting the target outcome (digit). Two possibilities exist: success or failure. The entries used for testing were not part of the training data set.

| algorithm id | errors | tests | error rate | accuracy |
|---|---|---|---|---|
| deodata_rasturnat_pow_e | 10877172 | 28569600 | 0.380725386 | 0.619274614 |
| deodata_tbreak_delanga | 10988837 | 28569600 | 0.384633912 | 0.615366088 |
| deodata_varsate_entropy | 11278683 | 28569600 | 0.394779171 | 0.605220829 |
| deodata_delanga | 11376191 | 28569600 | 0.398192169 | 0.601807831 |
| decision_tree_id3 | 12973421 | 28569600 | 0.454098797 | 0.545901203 |
| random_tree | 14184611 | 28569600 | 0.496493161 | 0.503506839 |
| uniform_random | 21427917 | 28569600 | 0.750025097 | 0.249974903 |

Table 1: Test results for prediction accuracy.

The algorithms used were:

- *deodata_delanga*
  The simplest implementation of the Proximity Concurrent Data Predictor variant.

- *deodata_tbreak_delanga*
  Similar to deodata_delanga but with an added tie breaking procedure.

- *deodata_varsate_entropy*
  An implementation of the Cascading Concurrent Data Predictor variant, using entropy as an impurity measure.

- *deodata_rasturnat_pow_e*
  An implementation of the Swapped Concurrent Data Predictor variant; the formula to update the outcome score is similar to the one shown in the appendix wherein the base of the exponential has been changed from two to "e" (Euler's number).

- *decision_tree_id3*
  An implementation of the ID3 algorithm with no pruning.

- *random_tree*
  A decision tree that randomly chooses the splitting attributes.

- *uniform_random*
  A classifier that picks possible outcomes in a uniform random manner. It is used only as a control indicator for the experiment.

Table 1 indicates that all proposed Concurrent Data Predictor methods perform better than the reference decision tree classifier. Also, as expected, the random tree classifier performs worse than the reference decision tree, yet still retains a considerable predictive power. The uniform random guess produces an error rate close to the expected theoretical rate.

Fig. 8 shows that a ranking in performance exists among the variants of the Concurrent Data

Predictor. In this test setup, the most accurate algorithm is "*deodata_rasturnat_pow_e*" followed in order by "*deodata_tbreak_delanga*", "*deodata_varsate_entropy*", and "*deodata_delanga*". The fact that "*deodata_tbreak_delanga*" performs better than the plain "*deodata_delanga*" was to be expected, as additional tie break processing is present in the former.

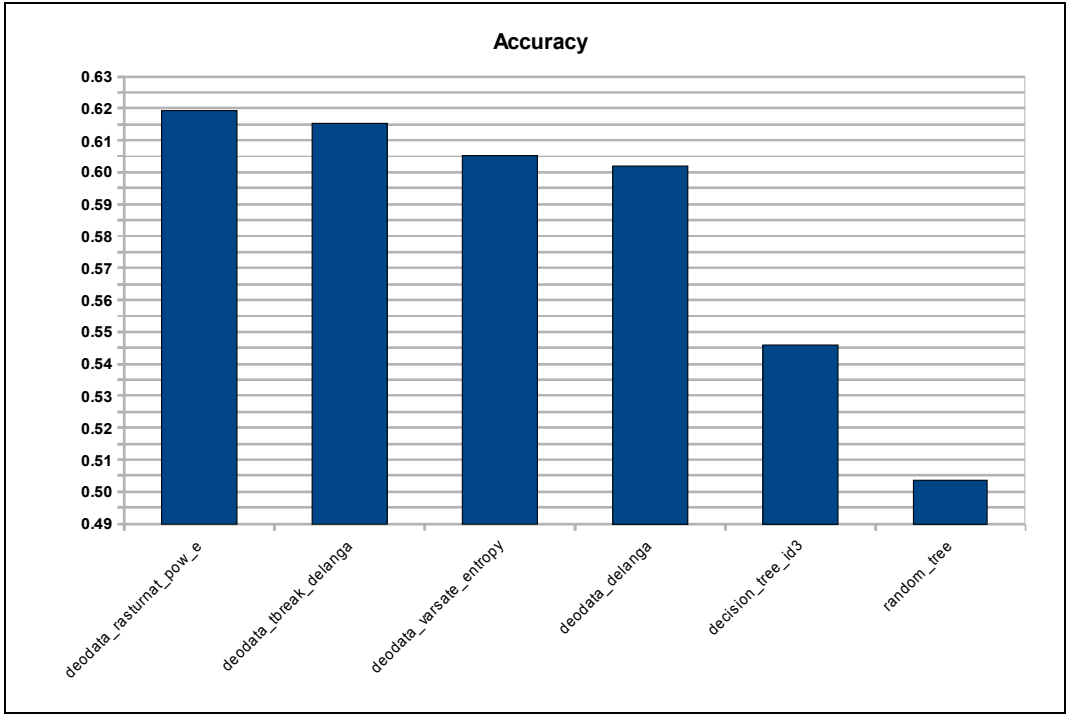

Figure 8: Comparison of test accuracy.

Table 2 contains the results of testing the evolution of the prediction accuracy as the training data set varies.

| per outcome train no | algorithm | | | |
|---|---|---|---|---|
| | deodata_delanga | decision_tree_id3 | random_tree | uniform_random |
| 1 | 0.538869848 | 0.523513665 | 0.506242159 | 0.333377016 |
| 2 | 0.585380824 | 0.579156586 | 0.562324149 | 0.333140681 |
| 4 | 0.623928091 | 0.620898297 | 0.610667563 | 0.334008737 |
| 8 | 0.655110887 | 0.654643817 | 0.648863127 | 0.334535170 |
| 16 | 0.675859095 | 0.675644041 | 0.672517921 | 0.333574149 |
| 32 | 0.689136425 | 0.688859767 | 0.687646729 | 0.332596326 |
| 64 | 0.697682572 | 0.697428315 | 0.697036290 | 0.332764337 |

Table 2: Test results, accuracy vs. training data size.

The results in Table 2 are for a prediction test with three possible target outcomes (digits) using four attributes (random pixels) with two possible values for pixel intensity ('on' or 'off'). The number of entries used in the training data set is variable, and each outcome has the same number of training examples in the data set.

Fig. 9 illustrates the evolution for the three selected algorithms:
- Proximity Concurrent Data Predictor (deodata_delanga)
- ID3 Decision Tree (decision_tree_id3)
- Random Tree (random_tree)

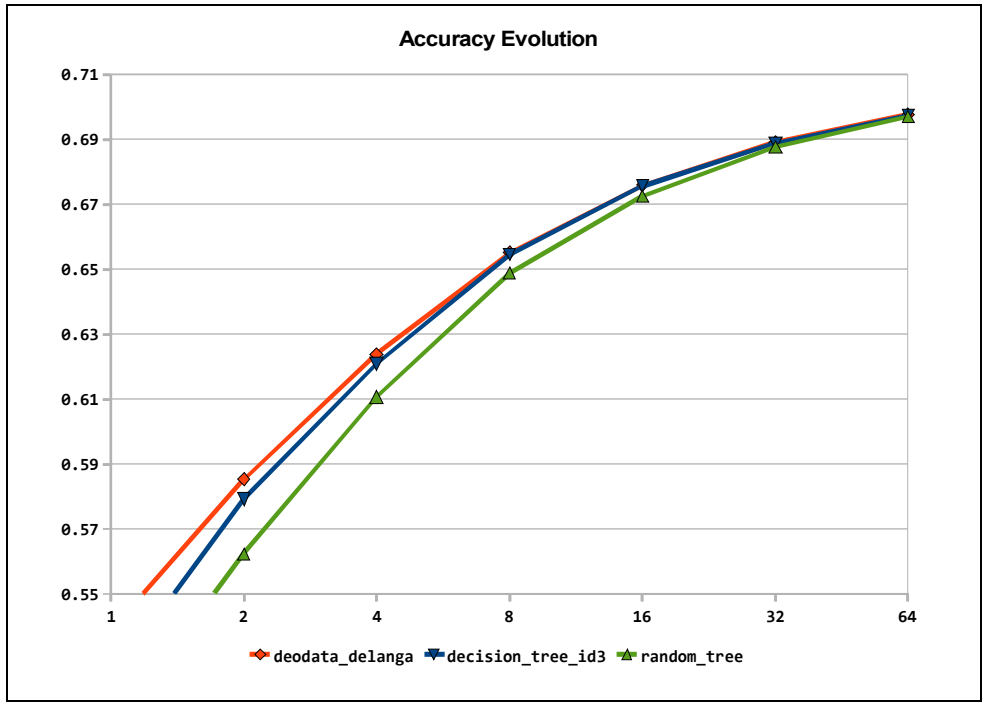

Figure 9: Test results, accuracy evolution.

The results suggest that the accuracies of the algorithms converge as the size of the training data set increases.

# Conclusions

A family of concurrent data predictors has been presented. It has been shown how the methods were derived from the decision tree classifier.

A converging equivalence between the standard decision tree and one of the variants has been described.

Experimental results have been presented suggesting the new methods improve the prediction accuracy over the reference decision tree algorithm.

# Appendix A. Code excerpts

In this section, pseudo code excerpts illustrate the operation of the presented methods.

Note that the excerpts use a pseudo programming language that imitates the style and structure of Python2. Some of the notable conventions of the language:

= : denotes assignment
== : denotes checking equality/equivalence
* : denotes multiplication, e. g. 2*3 == 6
** : denotes raising to the power, e. g. 2**3 == 8
# : precedes a comment

An example detailing the operation of the methods is provided..
The following simple training data is used for training:

```
#>-------------------------------------------------------------------
    # attr_table:
    #  the attribute table corresponding to the training data set
    attr_table == [
                    [ 'a1', 'b0', 'c2', 'd2', 'e2', 'f1' ],   # 't2'
                    [ 'a1', 'b0', 'c2', 'd0', 'e0', 'f1' ],   # 't1'
                    [ 'a1', 'b0', 'c1', 'd0', 'e0', 'f0' ],   # 't0'
                    [ 'a1', 'b0', 'c1', 'd1', 'e1', 'f2' ],   # 't2'
                    [ 'a1', 'b0', 'c1', 'd2', 'e2', 'f0' ],   # 't2'
                    [ 'a1', 'b1', 'c2', 'd0', 'e0', 'f1' ],   # 't1'
                    [ 'a1', 'b0', 'c2', 'd0', 'e0', 'f2' ],   # 't0'
                    [ 'a1', 'b0', 'c1', 'd0', 'e0', 'f2' ]    # 't1'
                  ]

    # targ_outc_list:
    #  the outcome list corresponding to the training data set
    targ_outc_list == ['t2', 't1', 't0', 't2', 't2', 't1', 't0', 't1']

#<-------------------------------------------------------------------
```

The query entry is the following:

```
#>-------------------------------------------------------------------
    query_attr = ['a1', 'b2', 'c1', 'd0', 'e1', 'f2']
#<-------------------------------------------------------------------
```

It has six attributes. The query entry is compared against each row of the training table.

Note that in this example the most simple score functions are used. For matching attribute values, a point is assigned in case of a match and none otherwise. The relevance/weight of the attribute, or the scarcity of the attribute values, are ignored in order to provide a simplified description.

```
#>----------------------------------------------------------------
    ###
    eval_outcome = 't2'
    eval_row =    ['a1', 'b0', 'c2', 'd2', 'e2', 'f1']
    query_attr = ['a1', 'b2', 'c1', 'd0', 'e1', 'f2']
    # matches        Y     N     N     N     N     N
    #        1 match
    col_score_list = [1, 0, 0, 0, 0, 0]
    # entry_match_score = EntryMatchEval(col_score_list)
    entry_match_score = sum(col_score_list)
    entry_match_score == 1

    if (operation_mode == 'delanga' or operation_mode == 'varsate') :
        work_dataset[entry_match_score] += eval_outcome
        work_dataset[1] == ['t2']
    else :
        # operation_mode == 'rasturnat'
        # transf_score = TransformScore(entry_match_score)
        transf_score = 2 ** entry_match_score
        work_dataset[eval_outcome] += transf_score
        work_dataset['t2'] == 2

    ###
    eval_outcome = 't1'
    eval_row =    ['a1', 'b0', 'c2', 'd0', 'e0', 'f1']
    query_attr = ['a1', 'b2', 'c1', 'd0', 'e1', 'f2']
    # matches        Y     N     N     Y     N     N
    #        2 matches
    col_score_list = [1, 0, 0, 1, 0, 0]
    entry_match_score = sum(col_score_list)
    entry_match_score == 2

    if (operation_mode == 'delanga' or operation_mode == 'varsate') :
        work_dataset[entry_match_score] += eval_outcome
        work_dataset[2] == ['t1']
    else :
        # operation_mode == 'rasturnat'
        transf_score = 2 ** entry_match_score
        work_dataset[eval_outcome] += transf_score
        work_dataset['t1'] == 4

    ###
    eval_outcome = 't0'
    eval_row =    ['a1', 'b0', 'c1', 'd0', 'e0', 'f0']
    query_attr = ['a1', 'b2', 'c1', 'd0', 'e1', 'f2']
    # matches        Y     N     Y     Y     N     N
    #        3 matches
    col_score_list = [1, 0, 1, 1, 0, 0]
    entry_match_score = sum(col_score_list)
    entry_match_score == 3

    if (operation_mode == 'delanga' or operation_mode == 'varsate') :
```

```
        work_dataset[entry_match_score] += eval_outcome
        work_dataset[3] == ['t0']
else :
        # operation_mode == 'rasturnat'
        transf_score = 2 ** entry_match_score
        work_dataset[eval_outcome] += transf_score
        work_dataset['t0'] == 8

###
eval_outcome = 't2'
eval_row =   ['a1', 'b0', 'c1', 'd1', 'e1', 'f2']
query_attr = ['a1', 'b2', 'c1', 'd0', 'e1', 'f2']
# matches      Y      N      Y      N      Y      Y
#      4 matches
col_score_list = [1, 0, 1, 0, 1, 1]
entry_match_score = sum(col_score_list)
entry_match_score == 4

if (operation_mode == 'delanga' or operation_mode == 'varsate') :
        work_dataset[entry_match_score] += eval_outcome
        work_dataset[4] == ['t2']
else :
        # operation_mode == 'rasturnat'
        transf_score = 2 ** entry_match_score
        work_dataset[eval_outcome] += transf_score
        work_dataset['t2'] == 2 + 16 == 18

###
eval_outcome = 't2'
eval_row =   ['a1', 'b0', 'c1', 'd2', 'e2', 'f0']
query_attr = ['a1', 'b2', 'c1', 'd0', 'e1', 'f2']
# matches      Y      N      Y      N      N      N
#      2 matches
col_score_list = [1, 0, 1, 0, 0, 0]
entry_match_score = sum(col_score_list)
entry_match_score == 2

if (operation_mode == 'delanga' or operation_mode == 'varsate') :
        work_dataset[entry_match_score] += eval_outcome
        work_dataset[2] == ['t1', 't2']
else :
        # operation_mode == 'rasturnat'
        transf_score = 2 ** entry_match_score
        work_dataset[eval_outcome] += transf_score
        work_dataset['t2'] == 18 + 4 == 22

###
eval_outcome = 't1'
eval_row =   ['a1', 'b1', 'c2', 'd0', 'e0', 'f1']
query_attr = ['a1', 'b2', 'c1', 'd0', 'e1', 'f2']
# matches      Y      N      N      Y      N      N
#      2 matches
col_score_list = [1, 0, 0, 1, 0, 0]
```

```
    entry_match_score = sum(col_score_list)
    entry_match_score == 2

    if (operation_mode == 'delanga' or operation_mode == 'varsate') :
        work_dataset[entry_match_score] += eval_outcome
        work_dataset[2] == ['t1', 't2', 't1']
    else :
        # operation_mode == 'rasturnat'
        transf_score = 2 ** entry_match_score
        work_dataset[eval_outcome] += transf_score
        work_dataset['t1'] == 4 + 4 == 8

    ###
    eval_outcome = 't0'
    eval_row =    ['a1', 'b0', 'c2', 'd0', 'e0', 'f2']
    query_attr = ['a1', 'b2', 'c1', 'd0', 'e1', 'f2']
    # matches       Y     N     N     Y     N     Y
    #        3 matches
    col_score_list = [1, 0, 0, 1, 0, 1]
    entry_match_score = sum(col_score_list)
    entry_match_score == 3

    if (operation_mode == 'delanga' or operation_mode == 'varsate') :
        work_dataset[entry_match_score] += eval_outcome
        work_dataset[3] == ['t0', 't0']
    else :
        # operation_mode == 'rasturnat'
        transf_score = 2 ** entry_match_score
        work_dataset[eval_outcome] += transf_score
        work_dataset['t0'] == 8 + 8 == 16

    ###
    eval_outcome = 't1'
    eval_row =    ['a1', 'b0', 'c1', 'd0', 'e0', 'f2']
    query_attr = ['a1', 'b2', 'c1', 'd0', 'e1', 'f2']
    # matches       Y     N     Y     Y     N     Y
    #        4 matches
    col_score_list = [1, 0, 1, 0, 1, 1]
    entry_match_score = sum(col_score_list)
    entry_match_score == 4

    if (operation_mode == 'delanga' or operation_mode == 'varsate') :
        work_dataset[entry_match_score] += eval_outcome
        work_dataset[4] == ['t2', 't1']
    else :
        # operation_mode == 'rasturnat'
        transf_score = 2 ** entry_match_score
        work_dataset[eval_outcome] += transf_score
        work_dataset['t1'] == 8 + 16 == 24
#<----------------------------------------------------------------------
```

After parsing all training entries, the work data set looks as follows:

```
#>--------------------------------------------------------------------
    if (operation_mode == 'delanga' or operation_mode == 'varsate') :
        work_dataset == {
                             1:['t2'],
                             2:['t1', 't2', 't1'],
                             3:['t0', 't0'],
                             4:['t2', 't1'],
                         }
    else :
        # operation_mode == 'rasturnat'
        work_dataset == {
                             't0': 16,
                             't2': 22,
                             't1': 24,
                         }
#<--------------------------------------------------------------------
```

The above descriptions are provided just as exemplifications to facilitate an understanding of the operating principle. Better implementations are possible.

The following pseudo-code excerpt illustrates how the work_dataset structure is processed in order to generate the likelihood measures:

```
#>--------------------------------------------------------------------
    ###
    if operation_mode == 'delanga' :
        ordered_score_list = GetScoreOrdered(work_dataset)
        ordered_score_list == [
                                  [4, ['t2', 't1']],
                                  [3, ['t0', 't0']],
                                  [2, ['t1', 't2', 't1']],
                                  [1, ['t2']]
                              ]
        # the top score corresponds to the first entry (index 0).
        top_score_list = ordered_score_list[0][1]
        top_score_list == ['t2', 't1']

        # the likelihood measure is represented by the outcome counts
        likelihood_data = GetCountToLikelihood(top_score_list)
        likelihood_data == [
                               {'outc':'t2', 'score':1},
                               {'outc':'t1', 'score':1}
                           ]

    ###
    elif operation_mode == 'varsate' :
        ordered_score_list = GetScoreOrdered(work_dataset)
        ordered_score_list == [
                                  [4, ['t2', 't1']],
```

```
                              [3, ['t0', 't0']],
                              [2, ['t1', 't2', 't1']],
                              [1, ['t2']]
                             ]

    # aggregate score lists into increasingly inclusive
    #   score lists starting from the top

    accumulated_list = []
    cascaded_list = []

    accumulated_list += ordered_score_list[0][1]
    accumulated_list == ['t2','t1']
    cascaded_list[0] = accumulated_list
    cascaded_list == [['t2','t1']]

    accumulated_list += ordered_score_list[1][1]
    accumulated_list == ['t2','t1','t0','t0']
    cascaded_list[1] = accumulated_list
    cascaded_list == [['t2','t1'], ['t2','t1','t0','t0']]

    accumulated_list += ordered_score_list[2][1]
    accumulated_list == ['t2','t1','t0','t0','t1','t2','t1']
    cascaded_list[2] = accumulated_list
    cascaded_list == [
                      ['t2','t1'],
                      ['t2','t1','t0','t0'],
                      ['t2','t1','t0','t0','t1','t2','t1']
                     ]

    accumulated_list += ordered_score_list[3][1]
    accumulated_list == ['t2','t1','t0','t0','t1','t2','t1','t2']
    cascaded_list[3] = accumulated_list
    cascaded_list == [
                      ['t2','t1'],
                      ['t2','t1','t0','t0'],
                      ['t2','t1','t0','t0','t1','t2','t1'],
                      ['t2','t1','t0','t0','t1','t2','t1','t2']
                     ]

    # evaluate the impurity score of each cascaded entry
    casc_score_list = []

    crt_ent = Entropy(cascaded_list[0])
    cascd_score_list[0] = [-crt_ent, cascaded_list[0]]
    crt_ent = Entropy(cascaded_list[1])
    cascd_score_list[1] = [-crt_ent, cascaded_list[1]]
    crt_ent = Entropy(cascaded_list[2])
    cascd_score_list[2] = [-crt_ent, cascaded_list[2]]
    crt_ent = Entropy(cascaded_list[3])
    cascd_score_list[3] = [-crt_ent, cascaded_list[3]]

    cascd_score_list == [
```

```
                       [-1.0,    ['t2','t1']],
                       [-1.5,    ['t2','t1','t0','t0']],
                       [-1.55665, ['t2','t1','t0','t0','t1','t2','t1']],
                       [-1.56127, ['t2','t1','t0','t0','t1','t2','t1','t2']]
               ]

       # the best score, in this mode of operation, corresponds to
       #   the lowest entropy, that is the first entry (index 0).
       top_score_list = ordered_score_list[0][1]
       top_score_list == ['t2', 't1']

       # the likelihood measure is represented by the outcome counts
       likelihood_data = GetCountToLikelihood(top_score_list)
       likelihood_data == [
                           {'outc':'t2', 'score':1},
                           {'outc':'t1', 'score':1}
                           ]
   ###
   else :
       # operation_mode == 'rasturnat'

       # order outcome entries using the accumulated score
       order_outc_list = OrderOutcomes(work_dataset)
       order_outc_list == [
                           [24, 't1'],
                           [22, 't2'],
                           [16, 't0']
                           ]
       # the likelihood measure is represented by the outcome score
       likelihood_data = GetScoreToLikelihood(order_outc_list)
       likelihood_data == [
                           {'outc':'t1', 'score':24}
                           {'outc':'t2', 'score':22},
                           {'outc':'t0', 'score':16}
                           ]
#<-------------------------------------------------------------------
```

It can be seen in the above example that, for operation modes 'delanga' and 'varsate', two outcome types have the same associated score. This is a tie situation. One possibility of breaking the tie is to search for additional counts in the below ensemble. The following pseudo-code excerpt illustrates such a procedure:

```
#>-------------------------------------------------------------------
    if operation_mode == 'delanga' :
        ordered_score_list = GetScoreOrdered(work_dataset)
        ordered_score_list == [
                               [4, ['t2', 't1']],
                               [3, ['t0', 't0']],
                               [2, ['t1', 't2', 't1']],
                               [1, ['t2']]
                               ]
```

```
        # the top score corresponds to the first entry (index 0).
        top_score_list = ordered_score_list[0][1]
        top_score_list == ['t2', 't1']

        # the likelihood measure is represented by the outcome counts
        likelihood_data = GetCountToLikelihood(top_score_list)
        likelihood_data == [
                            {'outc':'t2', 'score':1},
                            {'outc':'t1', 'score':1}
                            ]

        # to break the tie the next best entries are searched
        #    for additional counts:

        tie_index = 0
        next_entry_list = ordered_score_list[tie_index + 1][1]
        next_entry_list == ['t0', 't0']
        # the list doesn't contain instances of any of the tied
        #    outcomes 't2' or 't1', therefore it is ignored.

        # evaluate next index
        next_entry_list = ordered_score_list[tie_index + 2][1]
        next_entry_list == ['t1', 't2', 't1']
        # in this list there are 2 counts for 't1' and only one
        #    for 't2'. Therefore 't1' is chosen as the best prediction.

        predict_outcome = 't1'
#<-------------------------------------------------------------------
```